\documentclass[11pt]{article}

\usepackage[final]{acl}

\usepackage{times}
\usepackage{latexsym}

\usepackage[T1]{fontenc}

\usepackage[utf8]{inputenc}

\usepackage{microtype}

\usepackage{inconsolata}

\usepackage{graphicx}
\usepackage{booktabs}
\usepackage{amsmath}
\usepackage{hyperref}
\usepackage{cleveref}
\usepackage{multirow}
\usepackage[table]{xcolor}
\usepackage{makecell}

\title{Case2Flow: Bridging Patient Cases and Guideline Flowcharts through Multimodal Retrieval}

\author{
 \textbf{Jiale Wei\textsuperscript{1,2}},
 \textbf{Yufan Chen\textsuperscript{1}},
 \textbf{Alexander Jaus\textsuperscript{1,2}},
 \textbf{Zdravko Marinov\textsuperscript{1,2}},
\\
 \textbf{Julian Friedrich\textsuperscript{3}},
 \textbf{Simon Reiß\textsuperscript{1}},
 \textbf{Jens Kleesiek\textsuperscript{3}},
 \textbf{Rainer Stiefelhagen\textsuperscript{1}},
\\
\\
 \textsuperscript{1}Karlsruhe Institute of Technology, Karlsruhe, Germany
 \\
 \textsuperscript{2}Helmholtz Information and Data Science School for Health\\ (HIDSS4Health), Karlsruhe/Heidelberg, Germany
 \\
 \textsuperscript{3}University Hospital Essen, Essen, Germany
 \\
 \small{
   \textbf{Correspondence:} \href{mailto:jiale.wei@kit.edu}{jiale.wei@kit.edu}
 }
}

\begin{document}
\maketitle
\begin{abstract}
Medical guidelines encode rich, evidence-based decision logic, yet the specific decision artifact a clinician needs is hard to locate within a guideline, let alone across guidelines covering plausible diseases and treatments.
While guideline passages have supported end-to-end question answering, flowcharts remain largely underused in decision support despite their ability to encode actionable clinical pathways.
We therefore introduce \emph{Case2Flow}, a task designed to retrieve the most relevant guideline flowchart for a given patient case from a collection of guideline documents.
To support it, we construct \emph{FlowAtlas}, a curated corpus of $202$ flowcharts extracted from $2{,}080$ medical guidelines, together with a pipeline that synthesises $1{,}911$ aligned case-flowchart pairs.
Our evaluation of multimodal retrieval methods reveals systematic failure modes, including overreliance on keywords and spurious token-patch matches induced by uninformative background regions in flowcharts.
Motivated by this, we propose \emph{CRISP}, a training-free scoring method that sharpens late-interaction retrieval by suppressing uninformative patches, discounting ambiguous token matches, and incorporating bidirectional query-image alignment.
CRISP improves Recall@1 by up to $+18.71$~pp, while a blinded physician assessment on published case narratives provides preliminary feasibility evidence beyond synthetic queries.
Code and datasets are publicly available at \href{https://github.com/JialeWei/Case2Flow}{Case2Flow}.
\end{abstract}

\section{Introduction}
The volume of new medical guidelines is growing rapidly, with annual releases surging by up to $29\%$\footnote{\url{https://www.guidelinecentral.com/insights/2024-in-review-guidelines/}}, enriching evidence-based practice yet also increasing the burden on clinicians who must navigate and stay current with an ever-larger body of documents.
This makes it harder to put the decision support to practical use at the point of care.

One approach to making guideline knowledge accessible is training end-to-end language models such as BioBERT~\cite{lee2020biobert}, adapted for question answering, where guidelines serve as contextual inputs~\cite{li2025medguide} or training data~\cite{li2025rad,li2025medguide} to ground diagnoses and treatment decisions.
Retrieval-augmented methods~\cite{wang2026medagentpro,sohn2025rationale,staniek2025training} further link patient data with guideline passages.
However, these approaches operate on unstructured text and do not address visual decision artifacts within guidelines, such as flowcharts encoding clinical pathways.
Multimodal document retrieval has advanced through late-interaction models~\cite{faysse2025colpali,huang2025beyond}, and prior work on clinical flowcharts has focused on extracting decision logic~\cite{li2023meddm} or benchmarking LLM agents on predefined workflows~\cite{xiao2024flowbench}, but retrieving guideline flowcharts from patient cases remains unexplored.

\begin{figure*}[htbp]
    \centering
    \includegraphics[width=0.99\textwidth]{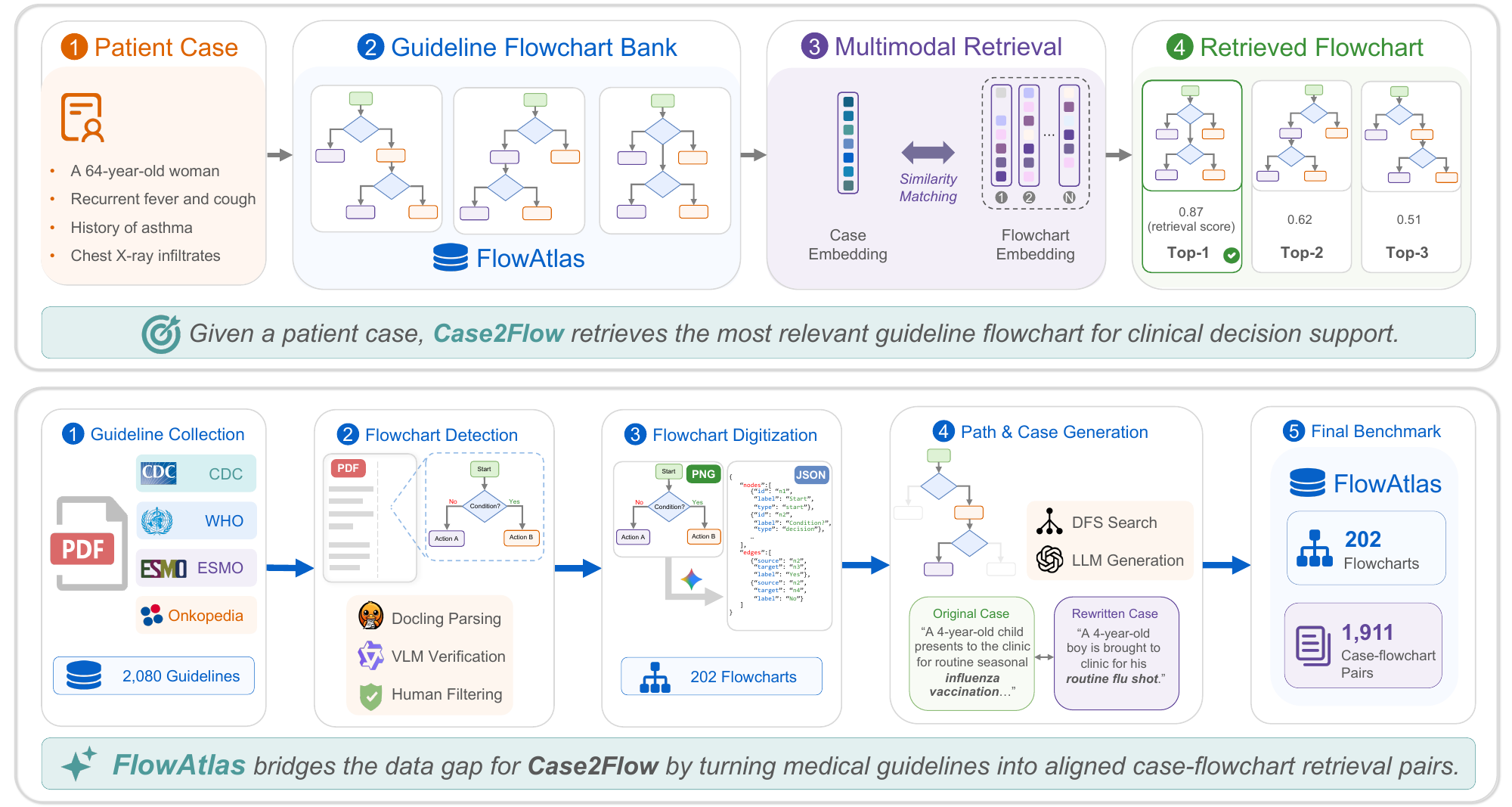}
    \caption{The task \emph{Case2Flow}: given a textual patient case (history, symptoms, key findings), retrieve the most relevant guideline flowchart from a large cross-provider candidate pool.
    We build \emph{FlowAtlas} from $2{,}080$ guidelines by extracting and digitising $202$ flowcharts and synthesising $1{,}911$ aligned case-flowchart pairs through DFS path enumeration, case generation, and clinical rewriting.}
    \label{fig:case2flow}
\end{figure*}

Rather than predicting a single \emph{`correct'} decision, an alternative is to retrieve the most relevant decision pathways for a given patient, letting practitioners make informed choices grounded in guideline logic.
We pursue this direction: specifically, we target the retrieval of flowcharts from medical guidelines, whose decision tree structures capture best practices while exposing multiple treatment pathways.
Our task \emph{Case2Flow}, shown in~\Cref{fig:case2flow}, aims to find the most relevant flowchart from a large set of medical guidelines, given a patient's clinical profile, including history, symptoms, and key findings.

To enable \emph{Case2Flow}, we first close the data gap with a synthetic generation pipeline that produces (\emph{textual patient case}, \emph{flowchart}) pairs.
We generate $1{,}911$ such pairs from $202$ flowcharts filtered from over two thousand candidate guidelines across four sources, spanning clinical contexts from treatment to prevention.
The resulting dataset \emph{FlowAtlas} allows us to benchmark multimodal retrieval systems on their ability to retrieve the source flowchart from a patient description.
We evaluate six methods on \emph{FlowAtlas}, investigate their failure modes, and also report results on the MedGUIDE dataset~\cite{li2025medguide}, originally proposed for question answering.
Finally, we show that shortcomings in the visual embeddings of late-interaction models can be addressed by our training-free \emph{CRISP} scoring, which suppresses uninformative image patches and verifies bidirectional query-image alignment.
Because the benchmark itself is synthetic, we additionally assess the retrieved charts on published clinical case narratives under blinded physician rating, which tests whether the task transfers beyond our generation pipeline.
Our contributions can be summarised as follows:
\begin{itemize}
    \item We define the novel retrieval task \emph{Case2Flow} (Case-based medical flowchart retrieval), which locates the decision artifact a clinician may wish to inspect.
    \item We close the data gap with a pipeline that extracts flowcharts from medical guidelines and produces flowchart-grounded synthetic patient cases, yielding \emph{FlowAtlas} with $1{,}911$ case-flowchart pairs.
    \item We benchmark existing retrieval methods on \emph{Case2Flow}, highlight current challenges, and show how our training-free \emph{CRISP} scoring improves Recall@1 by up to $+18.71$~pp.
\end{itemize}

\section{Related Work}
\textbf{Clinical decision support and guideline retrieval.}
Clinical decision support has gradually moved from closed-form prediction toward systems that ground recommendations in patient records, biomedical evidence, and clinical guidelines. Retrieval-augmented generation has been used to reduce hallucination and improve the use of evidence in medical question answering and guideline-based reasoning~\cite{wang2026medagentpro,sohn2025rationale,staniek2025training,li2025rad}. Recent oncology agents further show that language models can coordinate tools, knowledge bases, and guideline citations for complex treatment decisions~\cite{ferber2025development}. These systems demonstrate the value of connecting patient information to external clinical knowledge, but most of them retrieve textual passages, produce final answers, or assume that the relevant guideline context is already available. They therefore leave open a more basic access problem: given a patient case, how can a system locate the specific decision artifact that a clinician may wish to inspect?

\textbf{Multimodal document image retrieval.}
This access problem is closely related to document image retrieval, where recent work has moved beyond OCR pipelines toward direct visual retrieval. Benchmarks such as NL-DIR~\cite{guo2025towards} and MMDocIR~\cite{dong2025mmdocir} evaluate whether models can retrieve visually rich pages or layout elements from natural language queries. Visual RAG systems such as VisRAG~\cite{yu2025visrag} and VDocRAG~\cite{tanaka2025vdocrag} show that representing pages as images preserves layout, figures, tables, and charts that text extraction may weaken. Late-interaction models such as ColPali~\cite{faysse2025colpali} further improve retrieval by matching query tokens against visual patch representations rather than collapsing the page into a single embedding. These studies, however, are primarily designed for general documents or open-domain question answering. A clinical flowchart is more than a visually rich page: its relevance depends on whether a case satisfies its branching conditions and follows its pathway semantics.

\textbf{Flowchart understanding and retrieval.}
Prior work on medical flowcharts has mainly studied how to extract, formalise, or execute decision logic once the relevant source is known. MedDM~\cite{li2023meddm} and FT-MDT~\cite{li2025ft} convert medical text or guidelines into decision trees and executable guidance rules, while recent decision-tree extraction methods improve structured clinical rule construction~\cite{hou2025decision}. FlowBench~\cite{xiao2024flowbench} evaluates whether agents can follow workflow knowledge during planning, and MedGUIDE benchmarks whether large language models can make guideline-consistent decisions over curated NCCN decision-tree diagrams~\cite{li2025medguide}. These studies are complementary to ours, but they largely assume a selected workflow, a converted tree, or a predefined decision diagram. In practice, before a clinician or system can inspect a guideline pathway, the relevant visual artifact must first be located in a heterogeneous guideline collection. Case2Flow addresses this earlier access problem by retrieving the raw guideline flowchart that best matches an unstructured patient case. Our task therefore differs from executing known workflows or generating final clinical decisions, while preserving the visual form in which guideline logic is often published.

\section{Methodology}
\subsection{\emph{Case2Flow}: Case-based Flowchart Retrieval}
The \emph{Case2Flow} task is to identify the best-matching flowchart from a set of candidate flowcharts $\mathcal{F} = \{f_1, \dots, f_n\}$ given a textual patient case.
A query $c$ concatenates the case description with a focused clinical question; the reference answer is withheld from the retriever.
Retrieval is carried out by a scoring function $\theta$ that maps a query-flowchart pair to a real-valued score and should rank the target flowchart $f^\star \in \mathcal{F}$ first, i.e., $f^\star = \arg\max_{f_j \in \mathcal{F}} \theta(c, f_j)$.
We define the best match as the flowchart whose decision logic most directly applies to the patient's clinical situation.
For \emph{FlowAtlas}, $f^\star$ is the chart containing the decision path from which the case was synthesised, fixed before retrieval.
It is therefore a reproducible provenance target rather than a claim that no other chart in $\mathcal{F}$ could be clinically appropriate; \Cref{sec:clinical} examines the latter question directly.
MedGUIDE uses its own supplied case-chart mapping.
Obtaining such pairs in practice is costly, because experts must be familiar with every flowchart in $\mathcal{F}$, potentially beyond their specialty, to determine whether $(c, f_j)$ is the best match.
We address this data gap below.

\subsection{Building \emph{FlowAtlas}}
\label{sec:dataset_generation}

\paragraph{Guideline collection.}
We crawl $2{,}080$ guideline PDFs from four public providers chosen for both topical breadth and stylistic diversity:
CDC\footnote{\url{https://stacks.cdc.gov/}} (infectious disease, vaccination, public health),
WHO\footnote{\url{https://www.who.int/publications}} (global health, prevention),
ESMO\footnote{\url{https://www.esmo.org/guidelines}} (oncology),
and Onkopedia\footnote{\url{https://www.onkopedia-guidelines.info}} (haematology/oncology).
Per-source statistics and a stylistic comparison are provided in \Cref{tab:case_stats} and Appendix~\ref{sec:source_bias}.

\paragraph{Flowchart detection and digitisation.}
Candidate flowchart pages are first localised with Docling~\cite{Docling}, which combines caption keyword matching with its visual classifier.
A Qwen3-VL~\cite{bai2025qwen3} verifier then checks each candidate for clinical content, a valid structure of nodes and arrows, and branching logic. A final manual sweep removes residual false positives such as figures that are not decision charts or charts without a clear clinical purpose.
Each retained chart is digitised by Gemini-2.5-Flash into a structured graph: nodes (start, decision, action, terminal) carry text labels, and labelled edges encode branch conditions.

\paragraph{Path enumeration and case synthesis.}
From each flowchart graph, we enumerate start-to-terminal decision paths via DFS with cycle detection.
For each path, GPT-5-mini generates a synthetic patient case describing demographics, presenting complaints, and findings, using only information attached to nodes along the path; the prompt prohibits clinical details that do not appear in the flowchart.
The same pass also emits a focused clinical question about the next decision and the reference answer read off the terminal node of the path; only the case and the question enter the retrieval query.
A second LLM pass then rewrites the case and the question into natural clinical language and substitutes flowchart-specific phrasing with lay equivalents (e.g.\ \textit{``influenza vaccination''}~$\to$~\textit{``flu shot''}; \textit{``CD4 count below 200 cells/$\mu$L''}~$\to$~\textit{``severely immunodeficient''}), leaving the target flowchart and the answer unchanged.

\paragraph{Two evaluation modes.}
We retain both versions of each case and use them as parallel evaluation modes.
\texttt{Original} cases share lexical material with the source flowchart and measure retrieval accuracy when surface tokens are available as a cue.
\texttt{Rewritten} cases require recognition of the same clinical situation through paraphrases closer to bedside phrasing.
Reporting both modes distinguishes accuracy driven by lexical overlap from robustness to clinically equivalent phrasing.

\begin{figure*}[htbp]
    \centering
    \includegraphics[width=0.99\textwidth]{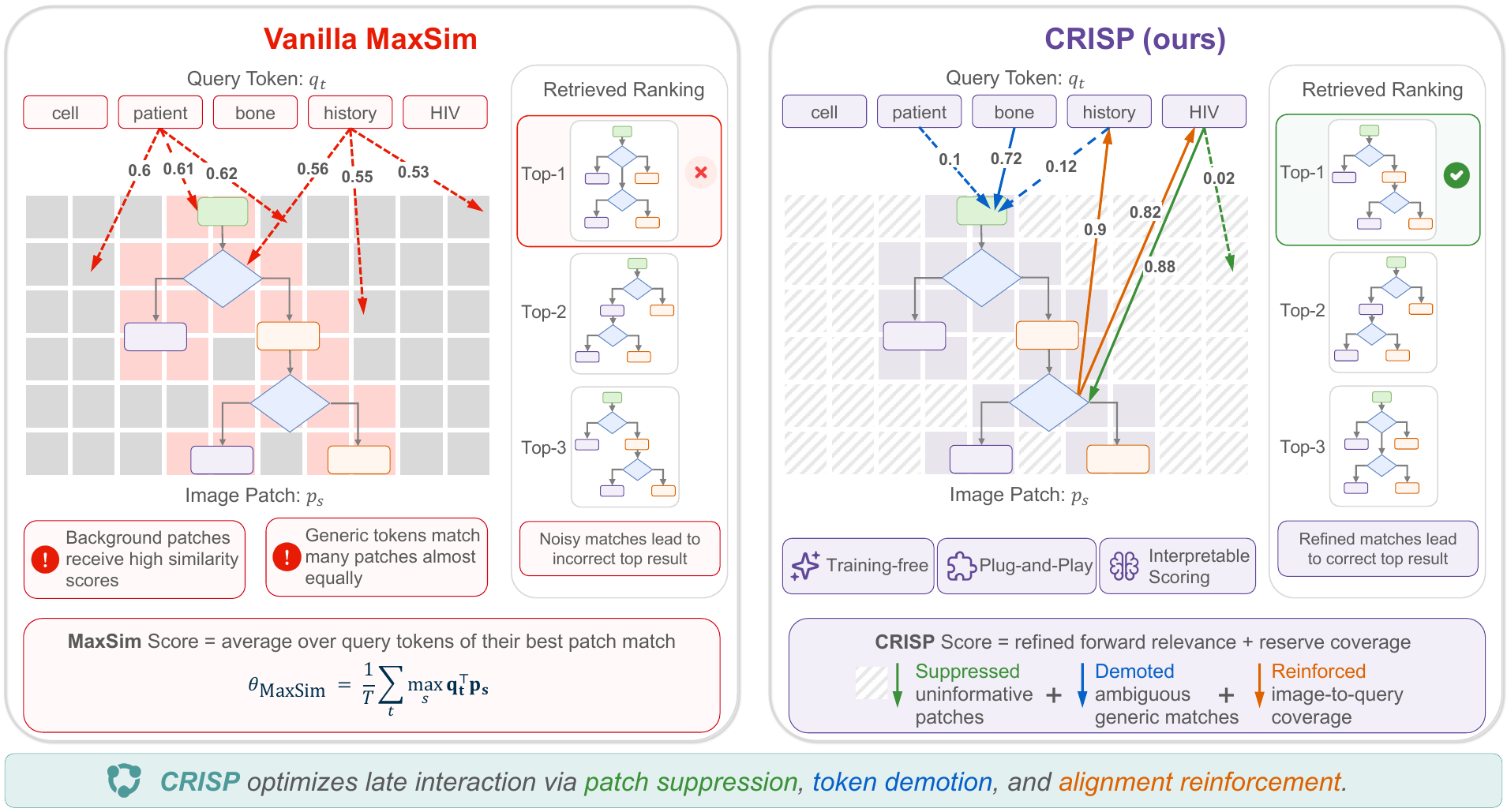}
    \caption{
    \emph{CRISP} sharpens late-interaction retrieval by reducing the influence of uninformative background patches, subtracting a soft uniform-match baseline from each query token to demote generic matches, and adding a reverse image-to-query coverage term.
    All three components are training-free and operate only at scoring time.
    }
    \label{fig:crisp}
\end{figure*}
\subsection{CRISP Scoring}
\label{sec:crisp}
Late-interaction VLMs such as ColPali~\cite{faysse2025colpali} and ColQwen3~\cite{huang2025beyond} encode a query $c$ into $T$ token embeddings $\mathbf{q}_1, \ldots, \mathbf{q}_T$ and a flowchart $f$ into $S$ per-patch embeddings $\mathbf{p}_1, \ldots, \mathbf{p}_S$, and score relevance by MaxSim:
$\mathrm{MaxSim}(c, f) = \frac{1}{T}\sum_{t=1}^{T} \max_{s \in [S]} \mathbf{q}_t^\top \mathbf{p}_s$.
Inspection of per-patch similarities on clinical flowcharts (cf.\ \Cref{fig:qualitative} and \Cref{fig:crisp}) reveals two recurring sources of error.
First, flowcharts contain substantial whitespace and background. Their patch embeddings cluster near a common background direction, yet their inner products with frequent query tokens can remain large enough for MaxSim to select them as best matches.
Second, frequent query tokens such as \textit{``patient''}, \textit{``history of''}, and \textit{``mg''} match many patches similarly. Because MaxSim takes the maximum for each token, these matches can contribute high scores that reflect token frequency rather than relevance.

\emph{CRISP} (\textbf{C}ross-directional \textbf{R}efined \textbf{I}nformative \textbf{S}coring of \textbf{P}atches) is a scoring modification that addresses both effects and adds a reverse-direction coverage term.
Let $\mathcal{A} = \{s \in [S] : \|\mathbf{p}_s\| > \varepsilon\}$ denote the active (non-padding) patches.
The full CRISP score combines three components, abbreviated \emph{PW}, \emph{Adj}, and \emph{Bi} below.

\paragraph{Patch weighting (PW).}
We compute the mean active embedding $\bar{\mathbf{p}} = \frac{1}{|\mathcal{A}|} \sum_{s \in \mathcal{A}} \mathbf{p}_s$ and weight each patch by its deviation from this mean:
\begin{equation}
w_s = \frac{d_s}{\sum_{s' \in \mathcal{A}} d_{s'}} \cdot |\mathcal{A}|,\;\; d_s = \|\mathbf{p}_s - \bar{\mathbf{p}}\|_2,
\label{eq:patch_weight}
\end{equation}
with $w_s = 0$ for $s \notin \mathcal{A}$.
Patches that carry text or decision-node content deviate from the average and receive higher weights; background patches that cluster near $\bar{\mathbf{p}}$ are suppressed.
The weights average to one over active patches, preserving the overall score magnitude.

\paragraph{Adjusted forward scoring (Adj).}
Instead of taking the single best weighted patch per token, we measure how much that best match stands out from the token's weighted average over all patches:
\begin{equation}
a_t = \max_{s \in \mathcal{A}}\, w_s \cdot \mathbf{q}_t^\top \mathbf{p}_s \;-\; \frac{1}{|\mathcal{A}|}\sum_{s \in \mathcal{A}} w_s \cdot \mathbf{q}_t^\top \mathbf{p}_s\,.
\label{eq:adjusted}
\end{equation}
A token with a sharp, distinctive patch match yields a large $a_t$; a token that matches everything near-uniformly produces $a_t \approx 0$ and contributes little to the score.

\paragraph{Bidirectional coverage (Bi).}
The forward term checks whether every query token finds a confident match in the flowchart, but it does not check whether the query addresses the flowchart content.
We therefore add a reverse MaxSim, denoted \emph{Bi}, in which each active patch selects its best query token, and we combine the two terms:
\begin{equation}
\theta_{\text{CRISP}}(c, f) = \frac{1}{T}\sum_{t=1}^{T} a_t \;+\; \frac{1}{|\mathcal{A}|}\sum_{s \in \mathcal{A}} \max_{t \in [T]}\, \mathbf{q}_t^\top \mathbf{p}_s\,.
\label{eq:crisp}
\end{equation}
The first term captures query-to-image precision; the second captures image-to-query coverage.
Normalising by $T$ and $|\mathcal{A}|$ keeps both on a comparable scale.
CRISP requires no training and can replace MaxSim in late-interaction retrieval models.

\section{Experiments}
\label{sec:results}

\begin{table*}[htbp]
    \centering
    \footnotesize
    \renewcommand{\arraystretch}{1.2}
    \setlength{\tabcolsep}{10pt}
    \begin{tabular}{@{} l c c cccccc}
        \toprule
        \textbf{Method} & \textbf{Size} & \textbf{Type} & \textbf{CDC} & \textbf{WHO} & \textbf{ESMO} & \textbf{Onko.} & \textbf{NCCN} & \textbf{Mixed} \\
        \midrule
        RANDOM & -- & -- & 7.69 & 1.92 & 1.12 & 2.08 & 1.82 & 0.39  \\
        \midrule
        
        \multicolumn{9}{c}{\textit{Textual Pipeline-Based Flowchart Retrieval}} \\
        \midrule
        
        \multirow{2}{*}{\makecell[l]{Tesseract OCR \\ + BM25}} & \multirow{2}{*}{--} & \texttt{Original} & \cellcolor{gray!10}83.82 & \cellcolor{gray!10}79.55 & \cellcolor{gray!10}77.65 & \cellcolor{gray!10}88.70 & \cellcolor{gray!10}50.59 & \cellcolor{gray!10}62.55 \\
         & & \texttt{Rewritten} & \cellcolor{gray!40}75.00 & \cellcolor{gray!40}72.16 & \cellcolor{gray!40}68.94 & \cellcolor{gray!40}64.81 & \cellcolor{gray!40} N/A & \cellcolor{gray!40}58.55 \\
        
        \multirow{2}{*}{\makecell[l]{VLM Captioning \\ + BM25}} & \multirow{2}{*}{2B} & \texttt{Original} & \cellcolor{gray!10}\textbf{89.71} & \cellcolor{gray!10}84.66 & \cellcolor{gray!10}80.56 & \cellcolor{gray!10}86.12 & \cellcolor{gray!10}71.28 & \cellcolor{gray!10}71.83 \\
         & & \texttt{Rewritten} & \cellcolor{gray!40}83.82 & \cellcolor{gray!40}76.42 & \cellcolor{gray!40}75.00 & \cellcolor{gray!40}67.38 & \cellcolor{gray!40}N/A & \cellcolor{gray!40}68.62 \\
        \midrule
        
        \multicolumn{9}{c}{\textit{Direct Image-Based Flowchart Retrieval}} \\
        \midrule
        
        \multirow{2}{*}{CLIP} & \multirow{2}{*}{0.4B} & \texttt{Original} & \cellcolor{gray!10}45.59 & \cellcolor{gray!10}50.57 & \cellcolor{gray!10}17.68 & \cellcolor{gray!10}13.16 & \cellcolor{gray!10}18.86 & \cellcolor{gray!10}15.17 \\
         & & \texttt{Rewritten} & \cellcolor{gray!40}45.59 & \cellcolor{gray!40}46.59 & \cellcolor{gray!40}13.76 & \cellcolor{gray!40}10.01 & \cellcolor{gray!40}N/A & \cellcolor{gray!40}14.73 \\

        \multirow{2}{*}{Qwen3-VL-Emb} & \multirow{2}{*}{8B} & \texttt{Original} & \cellcolor{gray!10}80.88 & \cellcolor{gray!10}85.80 & \cellcolor{gray!10}82.83 & \cellcolor{gray!10}92.70 & \cellcolor{gray!10}79.24 & \cellcolor{gray!10}74.39 \\
         & & \texttt{Rewritten} & \cellcolor{gray!40}76.47 & \cellcolor{gray!40}82.10 & \cellcolor{gray!40}79.17 & \cellcolor{gray!40}87.70 & \cellcolor{gray!40}N/A & \cellcolor{gray!40}73.05 \\
         
        \multirow{2}{*}{ColPali} & \multirow{2}{*}{3B} & \texttt{Original} & \cellcolor{gray!10}82.35 & \cellcolor{gray!10}86.08 & \cellcolor{gray!10}80.93 & \cellcolor{gray!10}80.40 & \cellcolor{gray!10}54.67 & \cellcolor{gray!10}55.48 \\
         & & \texttt{Rewritten} & \cellcolor{gray!40}82.35 & \cellcolor{gray!40}76.70 & \cellcolor{gray!40}68.81 & \cellcolor{gray!40}61.66 & \cellcolor{gray!40}N/A & \cellcolor{gray!40}52.09 \\

        \multirow{2}{*}{ColQwen3} & \multirow{2}{*}{4B} & \texttt{Original} & \cellcolor{gray!10}85.29 & \cellcolor{gray!10}86.65 & \cellcolor{gray!10}86.36 & \cellcolor{gray!10}96.14 & \cellcolor{gray!10}79.91 & \cellcolor{gray!10}78.44 \\
         & & \texttt{Rewritten} & \cellcolor{gray!40}83.82 & \cellcolor{gray!40}82.67 & \cellcolor{gray!40}82.95 & \cellcolor{gray!40}91.70 & \cellcolor{gray!40}N/A & \cellcolor{gray!40}76.93 \\

         \multirow{2}{*}{\textbf{\makecell[l]{ColPali \\ + \emph{CRISP} {\tiny(Ours)}}}} & \multirow{2}{*}{3B} & \texttt{Original} & \cellcolor{gray!10}85.29 & \cellcolor{gray!10}85.80 & \cellcolor{gray!10}82.07 & \cellcolor{gray!10}83.69 & \cellcolor{gray!10}73.38 & \cellcolor{gray!10}70.98 \\
         & & \texttt{Rewritten} & \cellcolor{gray!40}\textbf{85.29} & \cellcolor{gray!40}74.72 & \cellcolor{gray!40}69.44 & \cellcolor{gray!40}69.96 & \cellcolor{gray!40}N/A & \cellcolor{gray!40}68.23 \\
        
        \multirow{2}{*}{\textbf{\makecell[l]{ColQwen3 \\ + \emph{CRISP} {\tiny(Ours)}}}} & \multirow{2}{*}{4B} & \texttt{Original} & \cellcolor{gray!10}82.35 & \cellcolor{gray!10}\textbf{87.22} & \cellcolor{gray!10}\textbf{87.88} & \cellcolor{gray!10}\textbf{96.85} & \cellcolor{gray!10}\textbf{84.85} & \cellcolor{gray!10}\textbf{83.69} \\
         & & \texttt{Rewritten} & \cellcolor{gray!40}82.35 & \cellcolor{gray!40}\textbf{83.24} & \cellcolor{gray!40}\textbf{83.96} & \cellcolor{gray!40}\textbf{92.13} & \cellcolor{gray!40}N/A & \cellcolor{gray!40}\textbf{81.88} \\
        \bottomrule
    \end{tabular}
    \caption{Recall@1 (\%) across guideline sources and query types. \emph{Per-source}: candidate pool from one provider. \emph{NCCN}: external MedGUIDE~\cite{li2025medguide} flowcharts. \emph{Mixed}: all $257$ flowcharts pooled. MedGUIDE supplies its own case-chart pairs and has no rewriting pass, so the \texttt{Rewritten} row is not applicable (N/A) in the NCCN column. Best results per column in \textbf{bold}.}
    \label{tab:result}
\end{table*}

\subsection{Setup}
\label{sec:setup}
\paragraph{Settings.}
All models are evaluated zero-shot.
We report Recall@1 under three regimes:
(i) \emph{per-source}, where the candidate pool contains only flowcharts from a single provider and isolates in-domain retrieval;
(ii) \emph{mixed}, pooling all $202$ \emph{FlowAtlas} flowcharts with the $55$ flowcharts from MedGUIDE~\cite{li2025medguide} into $257$ candidates spanning four sources and an external benchmark style;
and (iii) \emph{external}, evaluating directly on the MedGUIDE pool with its $7{,}747$ cases, none generated by our pipeline.
In the \texttt{rewritten} Mixed setting the rewriting pass is applied only to the $1{,}911$ FlowAtlas queries; the $7{,}747$ MedGUIDE queries are external and carried over unchanged.
Recall@$\{1,3,5,10\}$ and MRR for all methods are reported in Appendix~\ref{sec:full_results}.

\paragraph{Baselines.}
We compare two families of methods.
\emph{Textual pipelines} extract text from each flowchart and retrieve with BM25: \textbf{OCR + BM25} (Tesseract) and \textbf{VLM Captioning + BM25} (Qwen3-VL-2B captions); stronger dense and hybrid variants are added in \Cref{tab:baselines}.
\emph{Direct image retrieval} computes cross-modal similarity directly: \textbf{CLIP}~\cite{radford2021learning}, \textbf{Qwen3-VL-Emb}~\cite{bai2025qwen3}, \textbf{ColPali}~\cite{faysse2025colpali}, and \textbf{ColQwen3}~\cite{huang2025beyond}.
For ColPali and ColQwen3 we additionally report \emph{CRISP} (\Cref{sec:crisp}); since it only changes the scoring function, its embeddings are identical to the corresponding baseline.

\subsection{Quantitative Results}

\Cref{tab:result} reports Recall@1 across all settings.
\paragraph{Per-source retrieval.}
Within a single provider, most methods clear $77\%$ on \texttt{original} cases, and ColQwen3 reaches $96.14\%$ on Onkopedia, indicating strong performance of modern late-interaction VLMs on familiar layouts.
CLIP is a clear outlier (e.g.\ $13.16\%$ on Onkopedia), reflecting how poorly its web caption-style pre-training transfers to clinical descriptions and flowcharts.

\paragraph{Original vs.\ rewritten cases.}
Across the $1{,}911$ pairs, the second LLM pass yields a mean token-set Jaccard similarity of $0.44$ between the two modes, with the substitutions concentrated on flowchart-specific vocabulary, and the two paradigms react differently.
\textit{Text pipelines decline substantially}: OCR + BM25 drops by $-23.89$~pp on Onkopedia and $-8.71$~pp on ESMO once lay vocabulary replaces flowchart phrasing.
\textit{Strong VLMs are more robust}: ColQwen3 and Qwen3-VL-Emb lose at most about $5$~pp per source.
ColPali sits between the two, in line with its smaller backbone, which is also where CRISP brings the largest gains.

\paragraph{External benchmark.}
On the external NCCN-based MedGUIDE dataset, ColQwen3 ($79.91\%$) and Qwen3-VL-Emb ($79.24\%$) lead, while the best textual pipeline, VLM captioning, reaches $71.28\%$. These trends mirror those on FlowAtlas, suggesting that the observed difficulty is not restricted to cases generated by our pipeline.

\paragraph{Cross-source mixed pool.}
With all $257$ flowcharts pooled, retrieval becomes substantially harder: ColQwen3 drops from $85$--$96\%$ per source to $78.44\%$, and ColPali from $80$--$86\%$ to $55.48\%$.
This setting is closest to deployment, where the relevant flowchart is rarely known to live in a single provider's pool.

\paragraph{CRISP.}
Applying CRISP as a drop-in scoring change improves both ColPali and ColQwen3 across nearly every setting, with the largest gains in the harder, larger-pool regimes.
On ColPali, mixed-pool Recall@1 improves by $+15.50$~pp ($55.48 \to 70.98$) and NCCN by $+18.71$~pp ($54.67 \to 73.38$).
On ColQwen3, mixed improves by $+5.25$~pp ($78.44 \to 83.69$) and NCCN by $+4.94$~pp ($79.91 \to 84.85$).
The larger improvement on ColPali is consistent with its weaker baseline being more affected by uniform weighting and background patches; ColQwen3 appears to mitigate part of this effect through stronger patch embeddings (cf.\ \Cref{tab:ablation}).
On the small CDC pool ($13$ candidates), CRISP slightly lowers ColQwen3 ($85.29 \to 82.35$): a net loss of two correct top-$1$ predictions over $68$ cases, better read as noise at this sample size than as a reliable regression, but marking the regime in which CRISP has least to offer.
Across all other settings, ColQwen3~+~CRISP achieves the best Recall@1, surpassing the $2\times$ larger Qwen3-VL-Emb.
A cluster bootstrap confirms that the ColQwen3 Recall@1 gains are significant in all three settings (Appendix~\ref{sec:stats}).
On Mixed, CRISP costs $1.26\times$ the scoring time on ColQwen3 and $1.49\times$ on ColPali, with under $0.2\%$ additional peak GPU memory.
\begin{table}[htbp]
    \centering
    \footnotesize
    \setlength{\tabcolsep}{6pt}
    \begin{tabular}{@{} l cc c @{}}
        \toprule
                        & \multicolumn{2}{c}{\textbf{Mixed}} & \textbf{NCCN} \\
        \cmidrule(lr){2-3}\cmidrule(lr){4-4}
        \textbf{Method} & \texttt{orig} & \texttt{rew} & \texttt{orig} \\
        \midrule
        OCR + BGE-M3                   & 56.74 & 54.23 & 57.45 \\
        OCR hybrid                     & 70.25 & 66.90 & 68.41 \\
        Graph-text + BGE-M3            & 55.50 & 53.46 & 59.06 \\
        Graph-text hybrid              & 67.60 & 65.82 & 70.12 \\
        \midrule
        OCR hybrid $\to$ BGE-reranker  & 51.34 & 49.53 & 54.30 \\
        OCR hybrid $+$ MaxSim          & 77.15 & 74.35 & 77.73 \\
        OCR hybrid $+$ \emph{CRISP}    & 80.48 & 77.68 & 81.18 \\
        \midrule
        ColQwen3 (MaxSim)              & 78.44 & 76.93 & 79.91 \\
        ColQwen3 $+$ \emph{CRISP}      & \textbf{83.69} & \textbf{81.88} & \textbf{84.85} \\
        \bottomrule
    \end{tabular}
    \caption{Recall@1 (\%) of stronger text, reranking, and fusion baselines on the full query sets. \emph{Hybrid}: equal-weight fusion of BM25 and BGE-M3 scores after per-query $z$-normalisation; \emph{graph-text}: the digitised flowchart graph serialised instead of OCR text.}
    \label{tab:baselines}
\end{table}

\paragraph{Alternative pipelines.}
Comparing visual retrieval against BM25 alone leaves open whether the gap reflects the modality or merely an under-tuned text pipeline, so \Cref{tab:baselines} varies both how the text is matched and what text is matched over.
Dense retrieval with BGE-M3~\cite{chen-etal-2024-m3} over the same OCR output is not by itself an improvement ($56.74$ against $62.55$ for BM25 on \texttt{original} Mixed), but their equal-weight hybrid is, gaining $7.7$ and $13.5$~pp over the two respectively and leading every other text-only configuration on Mixed.
Serialising the digitised flowchart graph instead of the OCR text trades one weakness for another, winning on NCCN ($70.12$ versus $68.41$) and losing on Mixed ($67.60$ versus $70.25$): the structure our pipeline recovers is not what the text channel is missing.
Taking the best text-only configuration in each setting, the gap to ColQwen3~+~CRISP still stands at $13.4$ to $15.0$~pp, so the multimodal advantage is not an artifact of a weak text baseline.

Two further ways of adding capacity on the text side fail as well.
A text cross-encoder (\texttt{bge-reranker-v2-m3})~\cite{chen-etal-2024-m3} applied zero-shot to the hybrid top-$10$ lowers Recall@1 by $14$ to $19$~pp while leaving Recall@10 unchanged by construction: the shortlist it receives already contains the correct chart in over $93\%$ of queries, and it reorders that shortlist worse than the first-stage scores do.
A domain mismatch is the likely cause, since the checkpoint is trained on natural-language passage pairs whereas flowchart OCR yields fragmentary node labels; we report it for this checkpoint and setup only, not for rerankers trained on the task.
Equal-weight score fusion of the OCR hybrid with ColQwen3 likewise degrades the visual scorer in all six settings, by $1.3$ to $2.6$~pp under MaxSim and by $3.2$ to $4.2$~pp under CRISP.
The larger degradation under CRISP is consistent with the rest of this section: sharpening the visual score leaves the text channel less to add.

A multimodal second stage, by contrast, does help, and is complementary to CRISP rather than an alternative to it.
Reranking the ColQwen3 top-$10$ with Qwen3-VL-Reranker-8B on a fixed sample of $500$ queries per setting, stratified by provider and initial-rank bucket, raises Recall@1 from $78.60$ to $91.20$ (\texttt{original} Mixed), $76.40$ to $90.00$ (\texttt{rewritten} Mixed), and $80.00$ to $94.60$ (NCCN).
These are paired with MaxSim on the identical sample and are not comparable to the full-set numbers in \Cref{tab:result,tab:baselines}; they show that a much larger cross-encoder recovers a large part of the residual error, at a cost far above that of a scoring-time change.

\subsection{Ablation}
\label{sec:ablation}
\Cref{tab:ablation} reports all combinations of the three CRISP components on MedGUIDE.
Each component is enabled on top of the MaxSim baseline, and we report Recall@1 and MRR on both backbones.
\emph{Adj} is ablated independently by setting all active-patch weights to one in Eq.~\eqref{eq:adjusted}, so no combination is tied to the presence of \emph{PW}.

\paragraph{Bi is the dominant single component.}
Adding \emph{Bi} alone yields the largest single-step gain on both backbones, with $+10.18$~pp R@1 on ColPali and $+2.68$~pp on ColQwen3.
This accounts for roughly half of the full CRISP improvement: $54\%$ of the R@1 lift on each backbone, and $56\%$ (ColPali) and $62\%$ (ColQwen3) of the MRR lift.
The image-to-query direction therefore provides the largest missing signal in vanilla MaxSim, and reverse MaxSim recovers part of it.

\paragraph{PW depends on the backbone's patch representations.}
\emph{PW} alone lifts ColPali by $+7.64$~pp R@1 but only $+0.86$~pp on ColQwen3.
ColQwen3 already $L_2$-normalises patch embeddings and explicitly masks non-image tokens during scoring, so the active-patch variation available to \emph{PW} is much smaller.
On the weaker, unmasked ColPali backbone, \emph{PW} down-weights background patches that MaxSim would otherwise score highly.

\paragraph{Max-minus-mean needs the reverse term.}
\emph{Adj} is not a standalone improvement: alone it gains $+3.26$~pp R@1 on ColPali but loses $3.61$~pp on ColQwen3, and \emph{PW\,+\,Adj} loses $4.67$~pp on ColQwen3 relative to the baseline.
Replacing per-token MaxSim with $\max - \mathrm{mean}$ removes absolute similarity magnitude in favour of contrast, which on its own is too noisy to rank flowcharts.
Reintroducing the reverse term restores the absolute-magnitude signal in the image-to-query direction, and \emph{Adj} then becomes the largest contributor: \emph{Adj\,+\,Bi} exceeds \emph{Bi} alone by $+7.22$~pp R@1 on ColPali and $+1.66$~pp on ColQwen3, and exceeds \emph{PW\,+\,Bi} on both.

\paragraph{The components interact rather than simply add.}
The full factorial shows the components to be complementary but not independently monotone.
On ColQwen3 the single-component gains sum to $-0.07$~pp R@1 while the full method reaches $+4.94$~pp; on ColPali the same combination is sub-additive ($+21.08$ summed versus $+18.71$ achieved), because \emph{PW} and \emph{Adj} partly address the same background-dominance effect, and once \emph{Adj\,+\,Bi} is in place \emph{PW} adds only $+1.31$ and $+0.60$~pp.
That the pattern holds across a weak and a strong backbone indicates that CRISP addresses a limitation of MaxSim aggregation rather than of a specific encoder.

\begin{table}[htbp]
    \centering
    \footnotesize
    \setlength{\tabcolsep}{3pt}
    \begin{tabular}{@{} l rr rr @{}}
        \toprule
                                  & \multicolumn{2}{c}{\textbf{ColPali}} & \multicolumn{2}{c}{\textbf{ColQwen3}} \\
        \cmidrule(lr){2-3}\cmidrule(lr){4-5}
        \textbf{Scoring variant}  & R@1 & MRR & R@1 & MRR \\
        \midrule
        MaxSim (baseline)             & 54.67 & 66.91 & 79.91 & 87.55 \\
        \;+ PW                        & 62.31 & 72.86 & 80.77 & 88.25 \\
        \;+ Adj                       & 57.93 & 68.70 & 76.30 & 84.49 \\
        \;+ Bi                        & 64.85 & 75.65 & 82.59 & 89.74 \\
        \;+ PW + Adj                  & 59.27 & 69.69 & 75.24 & 84.12 \\
        \;+ PW + Bi                   & 67.76 & 78.47 & 83.93 & 90.63 \\
        \;+ Adj + Bi                  & 72.07 & 81.16 & 84.25 & 90.58 \\
        \;+ \textbf{CRISP} (PW+Adj+Bi) & \textbf{73.38} & \textbf{82.55} & \textbf{84.85} & \textbf{91.08} \\
        \midrule
        $\Delta$ vs.\ MaxSim          & \emph{+18.71} & \emph{+15.64} & \emph{+4.94} & \emph{+3.53} \\
        \bottomrule
    \end{tabular}
    \caption{All combinations of the CRISP components on MedGUIDE ($7{,}747$ cases, $55$ flowcharts; Recall@1 and MRR in \%). \emph{PW}: patch weighting; \emph{Adj}: adjusted forward scoring; \emph{Bi}: bidirectional coverage.}
    \label{tab:ablation}
\end{table}

\subsection{Behavioural Analysis}
\label{sec:behavioural}

We examine two questions about CRISP's effect on the Mixed pool: how it changes the per-query rank of the ground truth, and which errors remain after it is applied (\Cref{tab:behavioural}).
\begin{table}[htbp]
    \centering
    \footnotesize
    \setlength{\tabcolsep}{7pt}
    \renewcommand{\arraystretch}{1.05}
    \begin{tabular}{@{} l cc cc @{}}
        \toprule
                                              & \multicolumn{2}{c}{\textbf{ColPali}} & \multicolumn{2}{c}{\textbf{ColQwen3}} \\
        \cmidrule(lr){2-3}\cmidrule(lr){4-5}
                                              & \texttt{orig} & \texttt{rew} & \texttt{orig} & \texttt{rew} \\
        \midrule
        \multicolumn{5}{@{}l}{\textit{Per-query rank change vs.\ MaxSim} (\%)} \\
        \;\;Improved                          & 37.4 & 39.9 & 14.5 & 15.2 \\
        \;\;Unchanged                         & 57.1 & 53.8 & 81.4 & 80.2 \\
        \;\;Hurt                              &  5.4 &  6.3 &  4.0 &  4.6 \\
        \;\;Recovered to rank $1$$^{\dagger}$ & 40.9 & 39.6 & 35.8 & 33.5 \\
        \midrule
        \multicolumn{5}{@{}l}{\textit{Top-$1$ errors remaining after CRISP}} \\
        \;\;Number of errors                  & 2{,}803 & 3{,}068 & 1{,}575 & 1{,}750 \\
        \;\;Same document (\%)                & 3.7  & 4.1  & 5.1  & 6.1  \\
        \;\;Same provider (\%)                & 51.4 & 49.4 & 67.4 & 62.2 \\
        \;\;Cross-provider (\%)               & 44.9 & 46.4 & 27.5 & 31.7 \\
        \bottomrule
    \end{tabular}
    \caption{Behavioural analysis of CRISP on the Mixed pool. \emph{Top}: per-query rank change of the ground-truth flowchart relative to MaxSim; $^{\dagger}$\,Recovery is a fraction of the queries MaxSim ranks below top-$1$, not of all queries. \emph{Bottom}: the top-$1$ errors remaining under CRISP, classified by the deepest level at which the prediction matches the ground truth.}
    \label{tab:behavioural}
\end{table}

\paragraph{Per-query rank changes.}
CRISP improves the rank of the ground-truth flowchart for $14.5\%$ to $39.9\%$ of queries and worsens it for only $4.0\%$ to $6.3\%$, depending on the backbone and mode. For most queries ($53.8\%$ to $81.4\%$), the rank is unchanged.
Among queries that MaxSim ranks below the first position, CRISP recovers $33.5\%$ to $40.9\%$ to rank $1$ across all four combinations of model and mode.
The mean gain when CRISP helps also exceeds the mean loss when it hurts ($+16$ versus $-2$ positions on ColPali, $+3$ versus $-2$ on ColQwen3), so CRISP concentrates its effect on difficult, incorrectly ranked queries rather than broadly reshuffling rankings.

\paragraph{Where the residual errors lie.}
When CRISP retrieves the wrong flowchart, we ask whether the prediction is clinically unrelated or a topical neighbour of the ground truth.
Every flowchart has three nested levels of identity: a \emph{provider}, a \emph{guideline document} within that provider's corpus, and a \emph{chart} within that document.
We partition each residual top-$1$ error by the deepest level at which the prediction matches the ground truth: \emph{same document} (a different chart in the same guideline, typically another stage or line of treatment), \emph{same provider} (a different document from the same body on a related subject), and \emph{cross-provider}.

\Cref{tab:behavioural} (bottom panel) reports this breakdown on the Mixed pool.
Among the $1{,}575$ ColQwen3+CRISP residual errors with \texttt{original} queries, only $27.5\%$ are cross-provider; $67.4\%$ retrieve a different guideline document on an overlapping topic and $5.1\%$ a different chart within the same document.
The same-provider class dominates every combination of model and mode and is concentrated in the NCCN subset, which contributes $95\%$ of these errors and contains many oncology flowcharts differing only in tumour subtype or staging.
Separately, a recurring CDC confusion occurs among three seasonal-influenza vaccination guidelines whose charts share most of their text.
ColPali shows the same pattern with more cross-provider errors ($44.9\%$), consistent with its weaker baseline ranking.
Most remaining errors are therefore topical neighbours rather than unrelated charts, matching the steep Recall@$k$ curve in Appendix~\ref{sec:full_results}: when the top-$1$ result is wrong, the correct chart is usually within a short shortlist.

\subsection{Clinical Validation on Published Cases}
\label{sec:clinical}
FlowAtlas measures whether a system recovers the chart a case was generated from, which is reproducible but not the same as whether that chart is clinically usable; we therefore also evaluate on real clinical narratives under a clinician's judgement.
\begin{table}[htbp]
    \centering
    \footnotesize
    \setlength{\tabcolsep}{5pt}
    \begin{tabular}{@{} l cc @{}}
        \toprule
        \textbf{Top-$1$ system} & \textbf{Accept.} ($\geq 2$) & \textbf{Applic.} ($=3$) \\
        \midrule
        ColPali (MaxSim)             & 9/50 \;\,($18\%$)  & 6/50 \;\,($12\%$)  \\
        \;+ \emph{CRISP}             & 13/50 ($26\%$)     & 6/50 \;\,($12\%$)  \\
        ColQwen3 (MaxSim)            & 22/50 ($44\%$)     & 14/50 ($28\%$)     \\
        \;+ \emph{CRISP}             & \textbf{24/50} ($48\%$) & 14/50 ($28\%$) \\
        \bottomrule
    \end{tabular}
    \caption{Blinded physician rating of the top-$1$ chart on $50$ published PMC-CaseReport narratives~\cite{wu2025towards}, scored $0$ (irrelevant) to $3$ (directly applicable).}
    \label{tab:clinical}
\end{table}

\paragraph{Protocol.}
We draw $50$ article-deduplicated case contexts from PMC-CaseReport~\cite{wu2025towards}, a corpus of published case reports.
Candidates are screened for library coverage by a text and title matching step independent of the four evaluated systems, and the sample is frozen before any system runs; the systems receive only the published case context.
The $133$ deduplicated case-chart pairs so returned are rated by one physician, blinded to system identity and to the screening information, from $0$ (irrelevant) to $3$ (directly applicable); multiple or no candidates may be acceptable.
\paragraph{Results.}
\Cref{tab:clinical} reports the outcome.
ColQwen3+CRISP returns an acceptable chart for $48\%$ of cases and a directly applicable one for $28\%$, ahead of both ColPali variants, and the ordering of the four systems matches their Recall@1 ordering on FlowAtlas.
Across the union of all returned candidates, $31$ of $50$ cases received at least one acceptable chart and $19$ received none.
Two limits apply: it rates only the returned candidates, so it measures what the systems surface rather than whether a suitable chart exists elsewhere in the library; and the MaxSim-CRISP differences are within rating noise here, so it supports the feasibility of the task rather than the specific gain from CRISP.

\subsection{Qualitative Results}
\begin{figure}[htbp]
    \centering
    \includegraphics[width=0.99\linewidth]{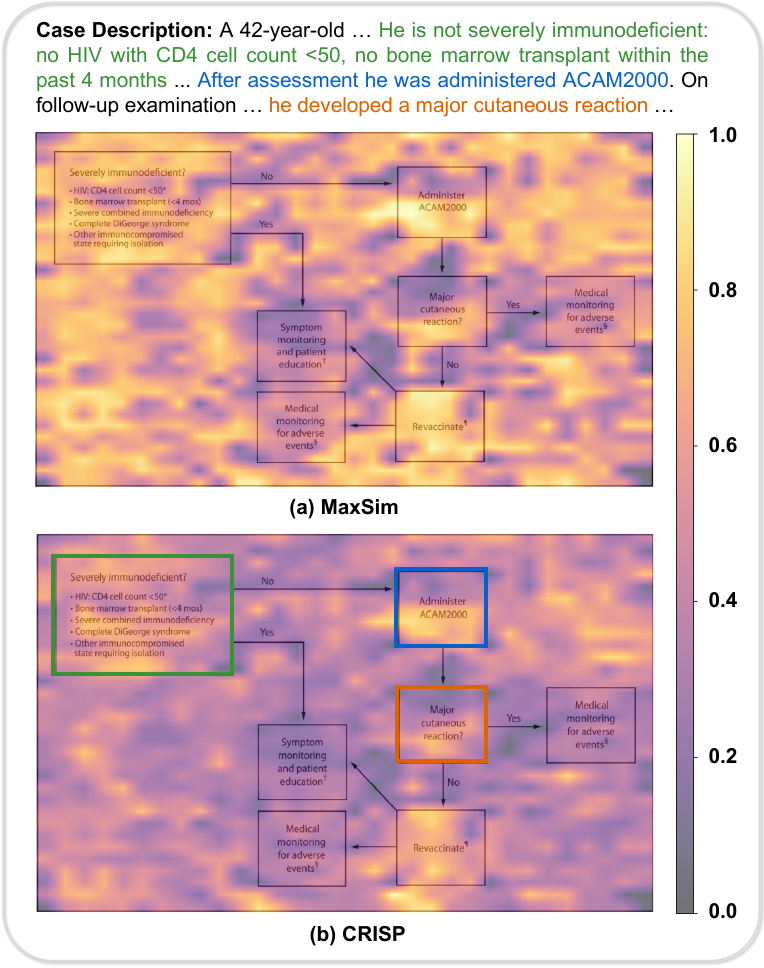}
    \caption{Per-patch similarity heatmaps on a CDC flowchart (ColPali). Top: standard MaxSim assigns high scores almost uniformly. Bottom: CRISP suppresses background and highlights decision nodes.}
    \label{fig:qualitative}
\end{figure}

\Cref{fig:qualitative} visualises per-patch similarity on a representative CDC flowchart with ColPali.
Under MaxSim, empty page background receives activations comparable to the decision nodes and a few generic tokens drive the score; CRISP suppresses those patches and concentrates similarity on decision nodes and clinical-term tokens, reflecting the two artifacts that motivated its design.
The pattern holds across providers, and Appendix~\ref{sec:qualitative} adds one heatmap per CRISP component.

\section{Conclusion}
We introduced \emph{Case2Flow}, the task of retrieving the guideline flowchart that applies to a patient case, and \emph{FlowAtlas}, $1{,}911$ case-flowchart pairs built over four guideline sources.
Benchmarking six retrieval methods exposed a failure mode of late-interaction scoring, in which uninformative background patches attract spurious matches; our training-free \emph{CRISP} scoring corrects it, improving Recall@1 by up to $+18.71$~pp at a small scoring-time cost.
A blinded physician assessment on published case narratives provides preliminary feasibility evidence beyond our synthetic queries, while leaving clear headroom for future work on real clinical notes.

\section*{Limitations}
\paragraph{Synthetic cases.}
FlowAtlas cases are LLM-generated from the labels along enumerated decision paths and then rewritten by a second LLM pass into more natural clinical phrasing.
The pipeline is grounded in the flowcharts and remains consistent with the encoded decision logic, but it does not reproduce the noise, ambiguity, and incomplete information of real clinical notes.
We mitigate this limitation in three ways: we report both \texttt{Original} and \texttt{Rewritten} modes, whose difference provides a proxy for paraphrase robustness; we evaluate on the external MedGUIDE benchmark, where all methods show trends consistent with those observed on FlowAtlas; and we assess retrieved charts on published clinical case narratives (\Cref{sec:clinical}).
That assessment has its own limits: a single rater, $50$ cases screened for library coverage and concentrated in oncology (no CDC or WHO targets), and labels on the returned candidates only.
It is feasibility evidence, not an unconditional estimate of real-world accuracy.
Validation on retrospective clinical notes requires governed clinical data and is beyond the scope of this study.

\paragraph{Closed-set evaluation.}
We evaluate retrieval on a closed pool of $202$ FlowAtlas flowcharts (and $55$ MedGUIDE flowcharts in the Mixed and external regimes).
This matches a deployment in which an institution curates and updates a guideline library, but it does not measure how methods scale to thousands of candidates or behave when no relevant chart exists in the pool.
CRISP's reverse-coverage term provides a per-query coverage-based score that could be evaluated for abstention, although we do not calibrate or evaluate that use here.

\paragraph{Scope of CRISP.}
CRISP is a scoring modification on top of late-interaction VLMs and does not alter the underlying embeddings, so the headroom it can recover is bounded by what the backbone already represents.
When the backbone fails to encode a piece of clinical content (for example, a dosage rendered as a small footnote), CRISP cannot recover it, because the relevant patch embedding lacks the information.
The small CDC regression on ColQwen3 (\Cref{sec:results}) is better treated as noise at this sample size; it occurs in the regime where CRISP has the least headroom---a small pool and a strong backbone.
As a practical rule, the settings in which CRISP helps most are those with many competing candidates and a backbone that leaves headroom; we do not have evidence for a pool-size threshold at which it should be disabled, and we do not propose an adaptive switch between MaxSim and CRISP.
CRISP is therefore complementary to backbone improvements rather than a replacement.

\paragraph{Source and language coverage.}
FlowAtlas covers four guideline providers and one external benchmark, all in English.
The four sources span layout style (text-heavy vs.\ diagrammatic), topical scope (public health vs.\ oncology), and provider type (national agency, international body, professional society, specialist consortium), but are not exhaustive.
Eligibility filtering also affects the providers very unevenly ($776 \to 13$ for CDC, $1{,}159 \to 52$ for WHO, $91 \to 89$ for ESMO, $54 \to 48$ for Onkopedia), so the corpus is concentrated in oncology: ESMO and Onkopedia together contribute $137$ of the $202$ charts.
National-specialty guidelines outside oncology and non-English guidelines are not represented.
The construction pipeline is provider-agnostic and released in full to facilitate extensions to further providers and languages.

\section*{Ethical Considerations}
\paragraph{Data and consent.}
FlowAtlas is built from publicly accessible guideline documents (CDC, WHO, ESMO, Onkopedia) and the publicly released MedGUIDE flowcharts~\cite{li2025medguide}.
The patient cases are synthesised from flowchart labels and contain no real patient records or personally identifying information.
The clinical validation in \Cref{sec:clinical} uses published case narratives from PMC-CaseReport~\cite{wu2025towards}; we redistribute case identifiers, not article content.

\paragraph{Human assessment.}
The blinded rating in \Cref{sec:clinical} was carried out by one clinician among the authors. No external annotators were recruited or compensated and no human-subject data were collected, so ethics board approval was not required.

\paragraph{Intended use and release.}
FlowAtlas and CRISP are released for research on case-based guideline retrieval.
Retrieving a flowchart is an access aid, not a clinical recommendation, and \Cref{sec:clinical} shows the top-$1$ chart to be unusable in roughly half of the assessed cases, so the resources are not suitable for automated clinical decision-making without expert oversight.
We release the pipeline, prompts, and code under an explicit licence with a per-component rights table, and provide reconstruction scripts where source documents may not be redistributed.

\section*{Acknowledgments}
The present contribution is supported by the Helmholtz Association under the joint research school “HIDSS4Health -- Helmholtz Information and Data Science School for Health”. This work was performed on the HoreKa supercomputer, funded by the Ministry of Science, Research and the Arts Baden-Württemberg and by the Federal Ministry of Education and Research.
\bibliography{ref}

\appendix
\clearpage
\section{FlowAtlas Statistics}
\label{sec:dataset_stats}
\Cref{tab:case_stats} reports per-source counts for \emph{FlowAtlas} and for the external MedGUIDE pool used in the \emph{Mixed} setting.
Path length denotes the number of decision nodes along the start-to-terminal path used to seed each synthetic case.
Orig.\,/\,Rew.\ Jaccard is the token-level Jaccard similarity between the two query modes, computed from sets of lower-cased alphabetic tokens; lower values indicate more surface rewriting.

\begin{table}[htbp]
    \centering
    \footnotesize
    \renewcommand{\arraystretch}{1.15}
    \setlength{\tabcolsep}{3pt}
    \begin{tabular}{@{} l r r r c c c @{}}
        \toprule
        \textbf{Source} & \textbf{Guide.} & \textbf{Flow.} & \textbf{Cases} & \textbf{Path} & \textbf{Words} & \textbf{Jacc.} \\
                        &                 &                &                & (avg)         & (o/r)          &                \\
        \midrule
        CDC       &   776 &  13 &    68 & 5.3 & 113/102 & 0.45 \\
        WHO       & 1{,}159 &  52 &   352 & 6.3 & 114/104 & 0.44 \\
        ESMO      &    91 &  89 &   792 & 5.8 & 103/103 & 0.44 \\
        Onkopedia &    54 &  48 &   699 & 7.7 & 106/105 & 0.45 \\
        \midrule
        \textbf{FlowAtlas} & \textbf{2{,}080} & \textbf{202} & \textbf{1{,}911} & 6.4 & 109/104 & 0.44 \\
        \midrule
        MedGUIDE  &   N/A &  55 & 7{,}747 & N/A & N/A     & N/A  \\
        \textbf{Mixed}     & N/A & \textbf{257} & \textbf{9{,}658} & N/A & N/A & N/A \\
        \bottomrule
    \end{tabular}
    \caption{FlowAtlas dataset statistics per guideline source. \emph{Guide.}: number of crawled guideline PDFs; \emph{Flow.}: retained flowcharts; \emph{Cases}: synthesised case-flowchart pairs; \emph{Path}: mean number of decision nodes per case; \emph{Words}: mean word count of (case~+~question) in \texttt{original}/\texttt{rewritten} mode; \emph{Jacc.}: mean token-level Jaccard similarity between the two modes. MedGUIDE is included for the Mixed and external evaluations but is not part of FlowAtlas.}
    \label{tab:case_stats}
\end{table}

\section{Per-Source Characteristics}
\label{sec:source_bias}
\begin{figure*}[htbp]
    \centering
    \includegraphics[width=0.99\textwidth]{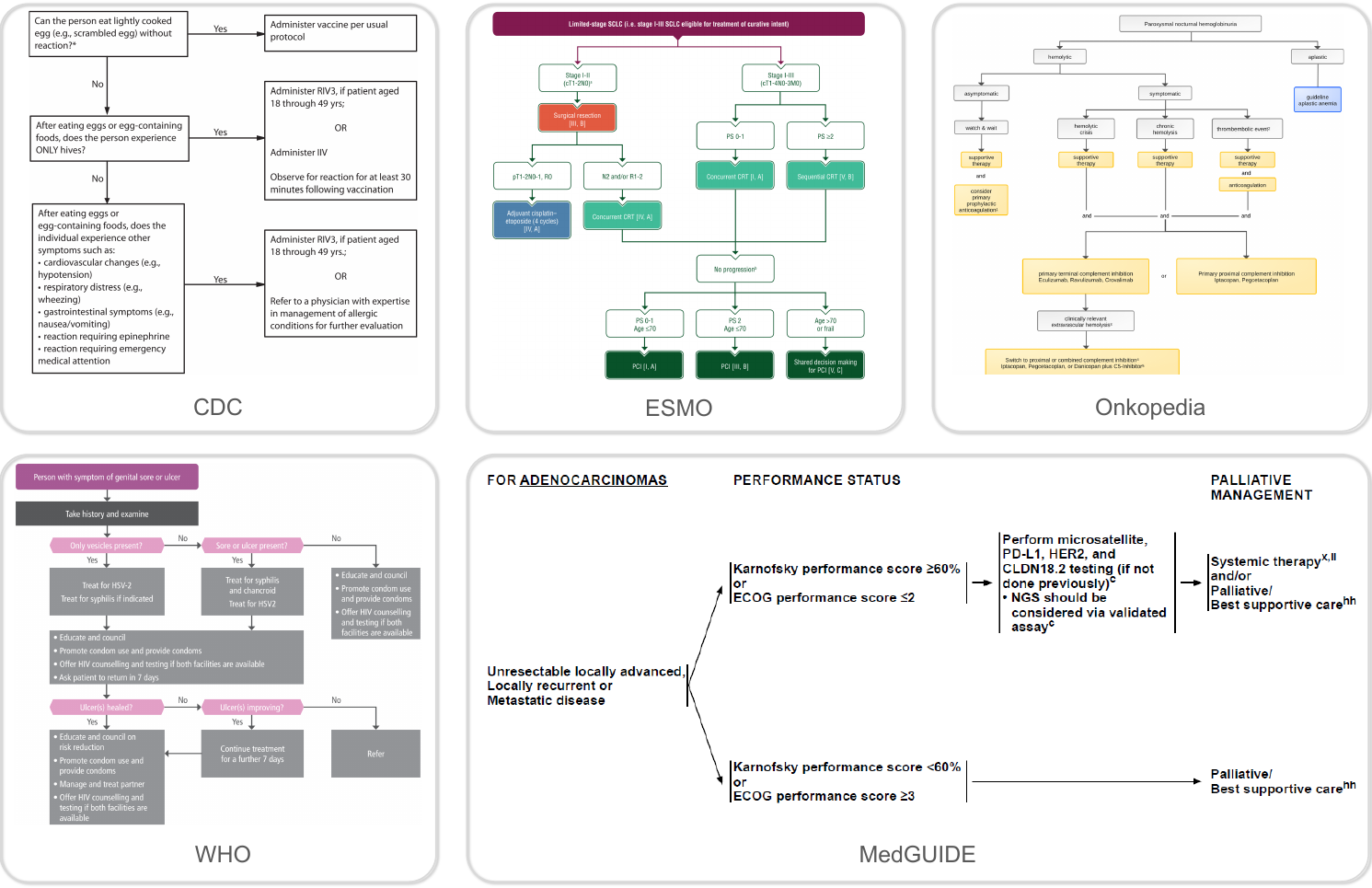}
    \caption{One representative flowchart from each of the five candidate pools (CDC, ESMO, Onkopedia, WHO and MedGUIDE).}
    \label{fig:flowchart_examples}
\end{figure*}

The four FlowAtlas sources differ in layout style, caption convention, and topic mix, and these differences affect retrieval difficulty.
\Cref{fig:flowchart_examples} shows one representative chart from each of the five candidate pools.
CDC is text-light and prescriptive. Its candidate pool is small and most methods approach saturation, leaving little room for CRISP to help.
WHO and ESMO cover heterogeneous topics with a mix of flow-diagram and table-like layouts; ColPali experiences its largest within-pool drops on these two sources.
Onkopedia is oncology-specialised and pharmacology-dense, and has the longest decision paths in FlowAtlas (up to $20$ nodes). ColQwen3+CRISP still reaches $96.85\%$ within this pool, suggesting that long, specialty-specific paths are not the primary limitation.
MedGUIDE is an external benchmark of cancer-staging flowcharts whose tabular layouts differ markedly from those of the four FlowAtlas sources; it is also where CRISP improves ColPali the most ($+18.71$~pp R@1).
The \emph{Mixed} pool combines all five sources and requires methods to discriminate among layout styles simultaneously. The spread between methods is widest in this setting, which most closely approximates cross-provider deployment.

\section{FlowAtlas Construction Pipeline}
\label{sec:dataset_gen_details}
\paragraph{Guideline collection.}
We collect $2{,}080$ guideline PDFs from four public providers (CDC, ESMO, Onkopedia, and WHO) and exclude non-English documents and non-guideline artifacts.
The four corpora vary in size and topical scope by design: CDC and WHO cover broad, population-oriented topics, whereas ESMO and Onkopedia are oncology-specialised.

\paragraph{Flowchart detection and digitisation.}
Candidate flowchart pages are first localised with Docling~\cite{Docling}, which combines caption keyword matching (\texttt{\{algorithm, flowchart, flow chart, pathway, decision tree, schema\}}) with its visual layout classifier.
Pages without a caption match but with a high visual-classifier score are also kept as candidates.
Each candidate is then passed to a Qwen3-VL-32B~\cite{bai2025qwen3} verifier with a structured three-criteria prompt: (i) clinical content, (ii) a valid structure of nodes and arrows, and (iii) at least one branch.
A final manual sweep by the authors removes residual false positives such as tabular figures, organisational charts, and charts without a clear clinical purpose.
The three-stage cascade yields $202$ retained flowcharts; most filtering occurs at the verification stage rather than during detection.
Each retained chart is digitised by Gemini-2.5-Flash into a JSON graph with typed nodes (\textit{start}, \textit{decision}, \textit{action}, \textit{terminal}) and labelled edges that carry the branch condition.
Charts whose digitised graph fails a connectivity check are discarded.

\paragraph{Path enumeration and case synthesis.}
We enumerate start-to-terminal decision paths in each graph via depth-first search with cycle detection.
Long charts can produce hundreds of paths; we cap the number per chart at $40$ and prefer paths that visit a previously unused node or edge in order to maintain case diversity.
For each path, GPT-5-mini writes a synthetic patient case strictly grounded in the labels along the path.
The system prompt restricts the model to demographics, history, presenting complaints, and findings that appear on the path; it prohibits clinical details not present in the chart (lab values, comorbidities, treatments).
A second pass then rewrites each case into natural clinical phrasing and substitutes flowchart vocabulary with lay equivalents (e.g., \textit{``influenza vaccination''}~$\to$~\textit{``flu shot''}; \textit{``CD4 count below 200 cells/$\mu$L''}~$\to$~\textit{``severely immunodeficient''}).
The two passes are kept separate so that the rewritten case preserves the same clinical situation while reducing measurable surface overlap (mean Jaccard $0.44$; \Cref{tab:case_stats}).

\paragraph{Post-generation filtering.}
Among the FlowAtlas cases we discard (i) empty or truncated outputs from either LLM pass, (ii) cases with fewer than $30$ tokens of clinical content, and (iii) cases whose rewritten and original versions are byte-identical, since these would collapse the two evaluation modes.
MedGUIDE cases are not passed through the rewriting pass and are therefore exempt from (iii).
The remaining $1{,}911$ pairs form the FlowAtlas evaluation set; per-source counts and word statistics are reported in \Cref{tab:case_stats}.

\paragraph{Post-generation validation.}
Manual review is applied at a single point in the pipeline: the sweep over detected flowchart pages described above, which decides chart eligibility.
Graph connectivity checking, path enumeration, case synthesis, and the post-generation filtering above are fully automated, and the generated cases were not clinically adjudicated.
Clinical judgement enters the study only in the separate assessment of \Cref{sec:clinical}.

\paragraph{Reproducibility.}
All four prompts (flowchart verification, graph extraction, case synthesis, and clinical rewriting) are released.
LLM calls use temperature $1.0$ for case generation (to encourage stylistic diversity) and temperature $0.3$ for verification and rewriting (where determinism matters more).
The rewriting pass also passes a fixed seed to the API, which makes regenerated cases close to the released ones but does not guarantee bit-exact reproduction.

\section{Per-Source and Mixed-Pool Results}
\label{sec:full_results}
For both query modes on the five candidate pools (four FlowAtlas sources and the external MedGUIDE pool), \Cref{tab:rk_persrc_a,tab:rk_persrc_b} report per-source Recall@$k$ ($k \in \{1, 3, 5, 10\}$) and MRR.
\Cref{tab:rk_mixed_app} provides the corresponding numbers on the \emph{Mixed} pool that combines all $257$ flowcharts.
Two patterns hold across the tables.
First, CRISP improves R@1 and MRR more than R@5 or R@10, sharpening the top of the ranking rather than reshuffling lower-ranked results.
Second, R@10 is already high for the strongest models (above $97\%$ for ColQwen3$+$CRISP on the Mixed pool in both query modes). The central challenge is therefore placing the correct flowchart at rank~$1$ rather than retrieving a relevant shortlist, which is precisely the objective that CRISP targets.

\begin{table*}[htbp]
    \centering
    \footnotesize
    \renewcommand{\arraystretch}{1.15}
    \setlength{\tabcolsep}{2pt}
    \begin{tabular}{@{} l ccccc ccccc ccccc @{}}
        \toprule
        & \multicolumn{5}{c}{\textbf{CDC}} & \multicolumn{5}{c}{\textbf{WHO}} & \multicolumn{5}{c}{\textbf{ESMO}} \\
        \cmidrule(lr){2-6}\cmidrule(lr){7-11}\cmidrule(lr){12-16}
        \textbf{Method} & R@1 & R@3 & R@5 & R@10 & MRR & R@1 & R@3 & R@5 & R@10 & MRR & R@1 & R@3 & R@5 & R@10 & MRR \\
        \midrule
        \multicolumn{16}{@{}l}{\textit{\texttt{Original} queries}} \\
        \midrule
        OCR + BM25            & 83.82 & 95.59 & 100.00 & 100.00 & 90.00 & 79.55 & 91.48 & 93.18 & 93.18 & 85.52 & 77.65 & 87.75 & 89.14 & 91.16 & 83.12 \\
        VLM Cap.\ + BM25      & 89.71 & 94.12 & 100.00 & 100.00 & 92.89 & 84.66 & 95.45 & 96.59 & 96.59 & 90.10 & 80.56 & 90.03 & 90.28 & 91.54 & 85.30 \\
        CLIP                  & 45.59 & 83.82 &  92.65 &  98.53 & 65.68 & 50.57 & 76.42 & 88.92 & 95.74 & 65.94 & 17.68 & 26.89 & 32.20 & 43.06 & 26.08 \\
        Qwen3-VL-Emb          & 80.88 & \textbf{98.53} & \textbf{100.00} & \textbf{100.00} & 89.58 & 85.80 & 95.17 & 96.59 & 96.59 & 90.47 & 82.83 & 91.54 & 92.42 & 93.43 & 87.38 \\
        ColPali               & 82.35 & 97.06 & \textbf{100.00} & \textbf{100.00} & 89.46 & 86.08 & 95.17 & 96.02 & 96.02 & 90.41 & 80.93 & 90.28 & 91.67 & 91.79 & 85.71 \\
        \;+ \emph{CRISP}      & 85.29 & 95.59 & \textbf{100.00} & \textbf{100.00} & 90.56 & 85.80 & \textbf{96.31} & \textbf{98.01} & \textbf{99.72} & 91.33 & 82.07 & 90.91 & 91.92 & 92.42 & 86.57 \\
        ColQwen3              & \textbf{85.29} & 97.06 & \textbf{100.00} & \textbf{100.00} & \textbf{91.18} & 86.65 & 95.17 & 96.59 & 96.59 & 90.94 & 86.36 & 91.92 & 92.93 & 94.07 & 89.59 \\
        \;+ \emph{CRISP}      & 82.35 & 94.12 & \textbf{100.00} & \textbf{100.00} & 88.97 & \textbf{87.22} & 95.17 & 96.59 & 96.59 & \textbf{91.37} & \textbf{87.88} & \textbf{94.07} & \textbf{94.19} & \textbf{94.32} & \textbf{91.09} \\
        \midrule
        \multicolumn{16}{@{}l}{\textit{\texttt{Rewritten} queries}} \\
        \midrule
        OCR + BM25            & 75.00 & 92.65 & 100.00 & 100.00 & 84.04 & 72.16 & 87.22 & 90.06 & 92.05 & 80.16 & 68.94 & 78.79 & 82.83 & 87.37 & 75.28 \\
        VLM Cap.\ + BM25      & 83.82 & 94.12 & 100.00 & 100.00 & 90.12 & 76.42 & 91.19 & 94.60 & 96.31 & 84.23 & 75.00 & 84.85 & 86.87 & 89.02 & 80.54 \\
        CLIP                  & 45.59 & 80.88 &  92.65 & 100.00 & 65.96 & 46.59 & 69.32 & 85.23 & 92.61 & 60.97 & 13.76 & 20.20 & 23.99 & 33.84 & 21.00 \\
        Qwen3-VL-Emb          & 76.47 & \textbf{97.06} & \textbf{100.00} & \textbf{100.00} & 86.52 & 82.10 & 94.89 & 96.59 & 97.44 & 88.52 & 79.17 & 91.04 & 92.42 & 93.56 & 85.17 \\
        ColPali               & 82.35 & \textbf{97.06} & \textbf{100.00} & \textbf{100.00} & 89.22 & 76.70 & 90.91 & 94.32 & 95.17 & 84.16 & 68.81 & 82.20 & 86.74 & 89.77 & 76.61 \\
        \;+ \emph{CRISP}      & \textbf{85.29} & 95.59 & \textbf{100.00} & \textbf{100.00} & \textbf{91.05} & 74.72 & 94.60 & \textbf{96.88} & \textbf{99.15} & 84.49 & 69.44 & 85.23 & 89.14 & 90.78 & 78.00 \\
        ColQwen3              & 83.82 & \textbf{97.06} & \textbf{100.00} & \textbf{100.00} & 90.44 & 82.67 & 94.03 & 96.02 & 96.31 & 88.21 & 82.95 & 91.54 & 91.92 & 93.43 & 87.58 \\
        \;+ \emph{CRISP}      & 82.35 & \textbf{97.06} & \textbf{100.00} & \textbf{100.00} & 89.22 & \textbf{83.24} & \textbf{94.32} & 96.31 & 96.59 & \textbf{88.76} & \textbf{83.96} & \textbf{93.56} & \textbf{93.81} & \textbf{94.19} & \textbf{88.72} \\
        \bottomrule
    \end{tabular}
    \caption{Per-source Recall@$\{1,3,5,10\}$ and MRR (\%) on CDC, WHO and ESMO. Best score per column among the direct-image rows in \textbf{bold}. Continued in \Cref{tab:rk_persrc_b}.}
    \label{tab:rk_persrc_a}
\end{table*}

\begin{table*}[tbp]
    \centering
    \footnotesize
    \renewcommand{\arraystretch}{1.15}
    \setlength{\tabcolsep}{8pt}
    \begin{tabular}{@{} l ccccc ccccc @{}}
        \toprule
        & \multicolumn{5}{c}{\textbf{Onkopedia}} & \multicolumn{5}{c}{\textbf{MedGUIDE}} \\
        \cmidrule(lr){2-6}\cmidrule(lr){7-11}
        \textbf{Method} & R@1 & R@3 & R@5 & R@10 & MRR & R@1 & R@3 & R@5 & R@10 & MRR \\
        \midrule
        \multicolumn{11}{@{}l}{\textit{\texttt{Original} queries}} \\
        \midrule
        OCR + BM25            & 88.70 & 93.85 & 96.42 &  98.14 & 91.92 & 50.59 & 68.40 & 76.20 & 88.00 & 62.49 \\
        VLM Cap.\ + BM25      & 86.12 & 92.27 & 93.56 &  95.28 & 89.69 & 71.28 & 89.14 & 92.89 & 96.48 & 81.02 \\
        CLIP                  & 13.16 & 21.89 & 28.18 &  37.63 & 22.29 & 18.86 & 35.52 & 46.44 & 62.68 & 32.46 \\
        Qwen3-VL-Emb          & 92.70 & 96.57 & 98.86 & \textbf{100.00} & 95.16 & 79.24 & 94.45 & 98.36 & 99.56 & 87.10 \\
        ColPali               & 80.40 & 87.84 & 90.56 &  92.85 & 85.19 & 54.67 & 74.62 & 82.66 & 89.14 & 66.91 \\
        \;+ \emph{CRISP}      & 83.69 & 90.56 & 93.28 &  97.14 & 88.13 & 73.38 & 90.59 & 93.93 & 97.17 & 82.55 \\
        ColQwen3              & 96.14 & 99.28 & 99.71 & \textbf{100.00} & 97.65 & 79.91 & 94.44 & 97.81 & 99.68 & 87.55 \\
        \;+ \emph{CRISP}      & \textbf{96.85} & \textbf{99.43} & \textbf{100.00} & \textbf{100.00} & \textbf{98.16} & \textbf{84.85} & \textbf{97.39} & \textbf{99.46} & \textbf{99.95} & \textbf{91.08} \\
        \midrule
        \multicolumn{11}{@{}l}{\textit{\texttt{Rewritten} queries}} \\
        \midrule
        OCR + BM25            & 64.81 & 79.40 & 85.41 &  91.42 & 73.83 & N/A & N/A & N/A & N/A & N/A \\
        VLM Cap.\ + BM25      & 67.38 & 85.26 & 88.84 &  92.70 & 77.27 & N/A & N/A & N/A & N/A & N/A \\
        CLIP                  & 10.01 & 19.31 & 23.89 &  33.91 & 19.06 & N/A & N/A & N/A & N/A & N/A \\
        Qwen3-VL-Emb          & 87.70 & 93.56 & 96.57 &  99.14 & 91.49 & N/A & N/A & N/A & N/A & N/A \\
        ColPali               & 61.66 & 72.53 & 76.11 &  80.26 & 68.76 & N/A & N/A & N/A & N/A & N/A \\
        \;+ \emph{CRISP}      & 69.96 & 76.11 & 78.68 &  84.98 & 75.05 & N/A & N/A & N/A & N/A & N/A \\
        ColQwen3              & 91.70 & 97.42 & 98.71 & \textbf{100.00} & 94.67 & N/A & N/A & N/A & N/A & N/A \\
        \;+ \emph{CRISP}      & \textbf{92.13} & \textbf{98.14} & \textbf{99.00} & \textbf{100.00} & \textbf{95.18} & N/A & N/A & N/A & N/A & N/A \\
        \bottomrule
    \end{tabular}
    \caption{Per-source Recall@$\{1,3,5,10\}$ and MRR (\%) on Onkopedia and the MedGUIDE pool. MedGUIDE does not provide separate rewritten cases, so its rewritten rows are marked N/A. Continued from \Cref{tab:rk_persrc_a}.}
    \label{tab:rk_persrc_b}
\end{table*}

\begin{table*}[htbp]
    \centering
    \footnotesize
    \setlength{\tabcolsep}{8pt}
    \begin{tabular}{@{} l ccccc ccccc @{}}
        \toprule
        & \multicolumn{5}{c}{\cellcolor{gray!10}\texttt{Original} \textbf{queries}} & \multicolumn{5}{c}{\cellcolor{gray!40}\texttt{Rewritten} \textbf{queries}} \\
        \cmidrule(lr){2-6}\cmidrule(lr){7-11}
        \textbf{Method} & R@1 & R@3 & R@5 & R@10 & MRR & R@1 & R@3 & R@5 & R@10 & MRR \\
        \midrule
        OCR + BM25            & 62.55 & 77.74 & 83.23 & 88.75 & 71.71 & 58.55 & 74.64 & 80.90 & 86.85 & 68.40 \\
        VLM Cap.\ + BM25      & 71.83 & 87.71 & 91.20 & 93.86 & 80.50 & 68.62 & 85.69 & 89.86 & 92.97 & 78.07 \\
        CLIP                  & 15.17 & 28.32 & 35.78 & 46.02 & 25.48 & 14.73 & 27.57 & 34.96 & 45.05 & 24.85 \\
        Qwen3-VL-Emb          & 74.39 & 90.33 & 93.95 & 96.75 & 82.90 & 73.05 & 89.98 & 93.69 & 96.57 & 82.03 \\
        ColPali               & 55.48 & 73.75 & 80.61 & 86.43 & 66.22 & 52.09 & 71.00 & 78.44 & 84.71 & 63.36 \\
        \;+ \emph{CRISP}      & 70.98 & 86.46 & 90.00 & 93.30 & 79.52 & 68.23 & 84.23 & 87.91 & 91.69 & 77.16 \\
        ColQwen3              & 78.44 & 91.83 & 95.15 & 97.91 & 85.72 & 76.93 & 91.26 & 94.68 & 97.64 & 84.68 \\
        \;+ \emph{CRISP}      & \textbf{83.69} & \textbf{95.30} & \textbf{97.62} & \textbf{99.01} & \textbf{89.75} & \textbf{81.88} & \textbf{94.64} & \textbf{97.25} & \textbf{98.85} & \textbf{88.56} \\
        \bottomrule
    \end{tabular}
    \caption{Recall@$\{1,3,5,10\}$ and MRR (\%) on the \emph{Mixed} pool ($257$ candidates, $9{,}658$ queries per mode).}
    \label{tab:rk_mixed_app}
\end{table*}

\section{Visualising the CRISP Score}
\label{sec:qualitative}
\begin{figure*}[htbp]
    \centering
    \includegraphics[width=\textwidth]{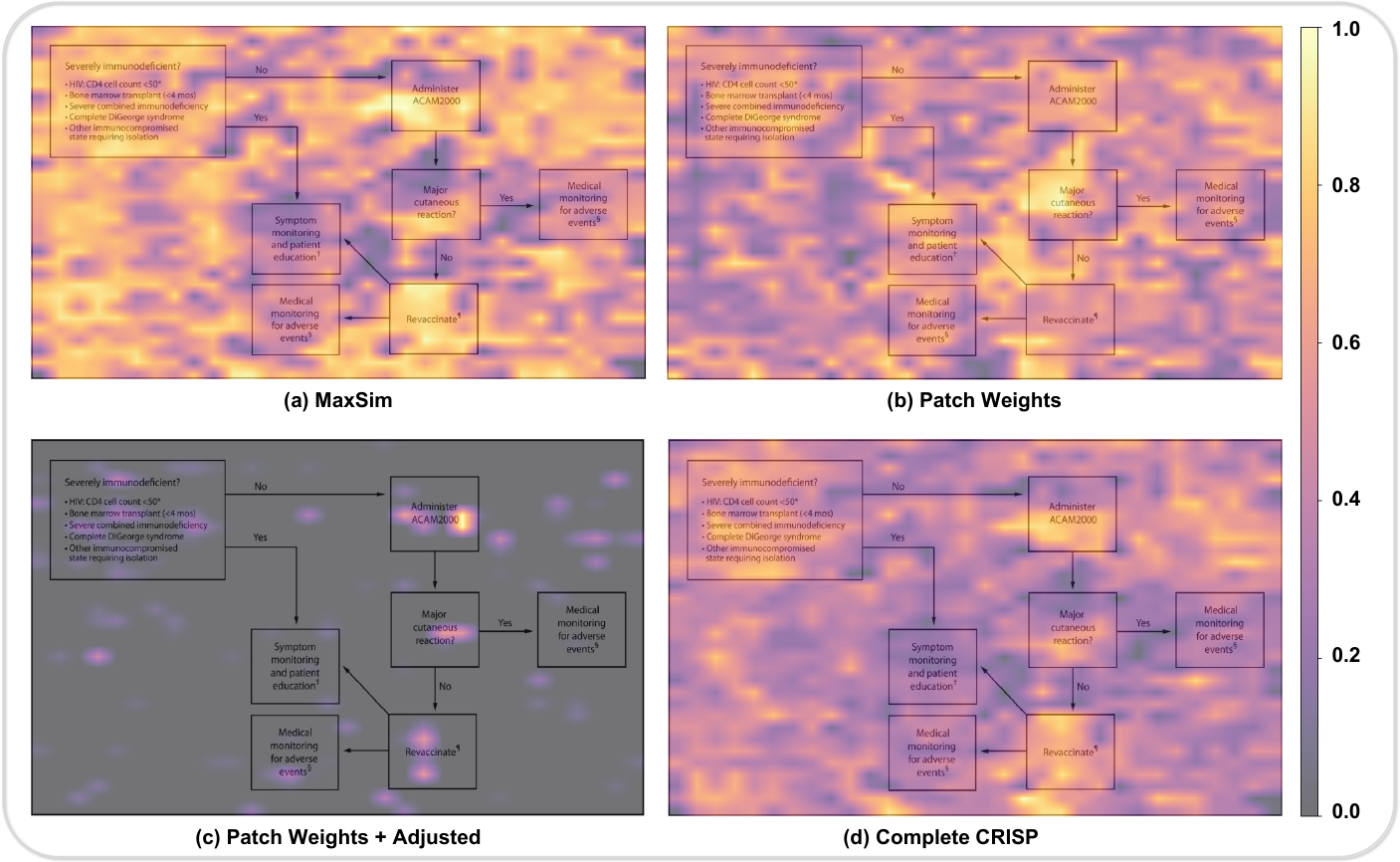}
    \caption{Per-patch decomposition of CRISP; each panel isolates one ingredient from the ablation table (\Cref{tab:ablation}).
    \textbf{(a) MaxSim}: $\max_t \mathbf{q}_t^\top \mathbf{p}_s$ saturates over whitespace.
    \textbf{(b) Patch weights}: $w_s$ from Eq.~\eqref{eq:patch_weight}; query-independent, concentrated on decision and outcome boxes.
    \textbf{(c) Patch weights + Adjusted}: per-token contrast $a_t$ from Eq.~\eqref{eq:adjusted} credited to its $\arg\max_s$ patch; sparse by construction.
    \textbf{(d) Complete CRISP}: image-side proxy $w_s \cdot \max_t \mathbf{q}_t^\top \mathbf{p}_s$ for Eq.~\eqref{eq:crisp}; the additive reverse term contributes a scalar per candidate with no separate per-patch signature.
    Each panel is normalised to its minimum and maximum values independently.}
    \label{fig:qual_ablation_cdc}
\end{figure*}

The main-text heatmap (\Cref{fig:qualitative}) contrasts MaxSim with CRISP, but it combines the three CRISP terms (\Cref{sec:crisp}) into a single ``before'' and ``after'' view.
\Cref{fig:qual_ablation_cdc} expands the comparison on the same chart and case by isolating one component at a time, allowing the contributions reported in the ablation table (\Cref{tab:ablation}) to be inspected on the patch grid.

\paragraph{(a) MaxSim.}
This panel renders $\max_t \mathbf{q}_t^\top \mathbf{p}_s$ on the image, the per-patch quantity that the token-side MaxSim baseline aggregates and that also appears as the (unweighted) second term of Eq.~\eqref{eq:crisp}.
On the CDC chart the map is bright across whitespace strips and connector arrows, with similarity scores comparable to those of the decision boxes themselves.
This is precisely the saturation failure mode that motivates the patch-weighting and adjusted-forward terms.

\paragraph{(b) Patch weights $w_s$.}
This panel renders the patch-weight map $w_s$ from Eq.~\eqref{eq:patch_weight}; it is query-independent and depends only on the $L_2$ distance between each patch embedding and the active-patch mean $\bar{\mathbf{p}}$.
High-weight regions trace the text-bearing decision and outcome nodes (\emph{Severely immunodeficient?}, \emph{Administer ACAM2000}, \emph{Major cutaneous reaction?}, \emph{Revaccinate}), whereas uniform whitespace lies near $\bar{\mathbf{p}}$ and receives little weight.
The contrast between (a) and (b) illustrates the geometric prior that PW contributes before any query token is matched, and it provides a spatial account of the $+7.64$~pp R@1 improvement attributed to PW on ColPali in \Cref{tab:ablation}.

\paragraph{(c) Patch weights $+$ adjusted forward.}
The adjusted-forward term in Eq.~\eqref{eq:adjusted} is per-token rather than per-patch: each query token $t$ contributes the contrast $a_t = \max_s w_s\, \mathbf{q}_t^\top \mathbf{p}_s - \frac{1}{|\mathcal{A}|}\sum_s w_s\, \mathbf{q}_t^\top \mathbf{p}_s$.
We render it spatially by attributing $a_t$ to the patch at which the maximum is achieved, so at most $T$ patches carry a non-zero value and the resulting map is sparse by construction.
The credited patches sit on the path-specific tokens that distinguish the case from generic clinical text: \emph{ACAM2000} in the administration node, \emph{cutaneous reaction} in the adverse-event branch, and the immunodeficiency-related tokens in the entry box.
On its own this term is too noisy to rank flowcharts (cf.\ \Cref{tab:ablation}); when combined with the reverse term, it contributes per-token contrast that PW alone cannot provide.

\paragraph{(d) Complete CRISP.}
The full score in Eq.~\eqref{eq:crisp} combines the per-token forward term in (c) with the unweighted reverse term in (a), each normalised by $T$ or $|\mathcal{A}|$.
For a single image-side proxy, we render $w_s \cdot \max_t \mathbf{q}_t^\top \mathbf{p}_s$, which keeps the same convention as (a) and (b) and shows how the weights from (b) reshape the coverage map from (a): whitespace activations are suppressed while decision-node activations are preserved.
The closed-form score additionally accumulates the per-token contrast in (c), but that contribution is a scalar per candidate and has no per-patch signature of its own.

\section{Statistical Reliability and Cost}
\label{sec:stats}
\paragraph{Significance.}
Queries are not independent, since many are generated from different decision paths through the same flowchart, so we resample \emph{ground-truth flowcharts} as clusters ($255$ on Mixed, $53$ on NCCN) in a paired bootstrap over $10{,}000$ replicates.
The $95\%$ intervals for the ColQwen3 Recall@1 improvement are $[2.35, 8.29]$, $[2.03, 8.07]$, and $[1.24, 8.85]$~pp on \texttt{original} Mixed, \texttt{rewritten} Mixed, and NCCN.
None includes zero; an exact paired McNemar test gives $p < 10^{-40}$ in all three, with CRISP correcting $602$--$747$ queries that MaxSim ranks wrongly against $220$--$269$ in the opposite direction.

\paragraph{Cost.}
CRISP reuses the backbone embeddings, so its cost is confined to scoring.
On one RTX~2080~Ti, scoring $9{,}658$ Mixed queries against $257$ candidates from cached embeddings takes $101.2$ versus $151.0$\,s for ColPali ($1.49\times$, $+5.2$\,ms per query) and $222.3$ versus $280.0$\,s for ColQwen3 ($1.26\times$, $+6.0$\,ms), with peak GPU memory rising by under $0.2\%$ ($266.2 \to 266.7$ and $355.2 \to 355.7$\,MiB).
Embedding generation, OCR, and model loading are identical under both scoring functions and excluded.

\end{document}